# On Calibration of Three-axis Magnetometer

Yuanxin Wu and Wei Shi

**Abstract**—Magnetometer has received wide applications in attitude determination and scientific measurements. Calibration is an important step for any practical magnetometer use. The most popular three-axis magnetometer calibration methods are attitude-independent and have been founded on an approximate maximum likelihood estimation (ML) with a quartic subjective function, derived from the fact that the magnitude of the calibrated measurements should be constant in a homogeneous magnetic field. This paper highlights the shortcomings of those popular methods and proposes to use the quadratic optimal ML estimation instead for magnetometer calibration. Simulation and test results show that the optimal ML calibration is superior to the approximate ML methods for magnetometer calibration in both accuracy and stability, especially for those situations without sufficient attitude excitation. The significant benefits deserve the moderately increased computation burden. The main conclusion obtained in the context of magnetometer in this paper is potentially applicable to various kinds of three-axis sensors.

## I. INTRODUCTION

Magnetometers are typically used for attitude determination and scientific measurements [1]. A three-axis magnetometer can measure the external geomagnetic field from which the local north direction can be derived, so it is frequently used to assist low-cost inertial measurement units to provide orientation information with bounded errors. Magnetometer is prone to the magnetic disturbance in the surrounding environment, such as the ferromagnetic material and strong electric currents. When the magnetometer is placed rigidly on or near to ferromagnetic objects, its output is distorted and cannot measure the external magnetic field. The distortion can be divided into hard iron and soft iron effects. The hard iron effect is simply the additive magnetic field produced by permanent magnets or electrical currents, while the soft iron effect is induced by materials that generate their own magnetic field in response to and distort the underlying magnetic field in both intensity and orientation. The three-axis magnetometer also exhibits scale factor, cross-coupling and bias errors, but these errors behave in the same manner with and are not discriminable from the soft/hard iron effects [2]. Careful calibration might be necessary each time the magnetometer is used.

Classical magnetometer calibration techniques (like the swing method [3]) require levelling and external known heading sources. However, end users prefer to an in-situ calibration with no requirement of external equipment. This practical demand gives birth to a class of attitude-independent calibration methods [4-11], which was first proposed in public literature by [7, 8] and has become popular in the last decade. These methods exploit the fact that the magnitude of magnetometer measurements is constant regardless of the orientation at the local position. The idea has also been applied to calibrate inertial sensors like accelerometers and gyroscopes [6, 9, 12-15]. The constant magnitude relationship is usually employed to estimate the calibration parameters in the form of the maximum likelihood (ML) estimation problem. The works [7, 8] pose the magnetometer calibration as an approximate ML problem and introduce a centering approximation technique to generate a good initial estimate for solving the resultant quartic objective function. A recursive calibration method based on Kalman filtering is proposed in [16] using the constant magnitude relationship as a pseudo measurement model. A simpler linearized batch least-square estimation is given in [3], in which the initial estimate is obtained by a pseudo-linear equation in intermediate variables. The work [4] claims that the magnetometer calibration is equivalent to the ML estimation on the ellipsoid manifold and uses the Gauss-Newton method to solve the approximate ML estimation. Therein another optimal ML estimation conditioned on auxiliary magnetic vectors in addition to the ellipsoid manifold is touched upon, but has not been actually implemented

---

This work was supported in part by the Fok Ying Tung Foundation (131061) and National Natural Science Foundation of China (61174002, 61422311).

Yuanxin Wu (corresponding author), School of Electronic Information and Electrical Engineering, Shanghai Jiao Tong University, Shanghai, China, 200240, E-mail: (yuanx_wu@hotmail.com).

Wei Shi, School of Aeronautics and Astronautics, Central South University, Changsha, Hunan, China, 410083. Tel: 086-0731-88877132, E-mail: (ahshw@csu.edu.cn).

due to the considerably enlarged parameter dimension incurred by the unavoidable auxiliary magnetic vectors. Similar ML formulation is used in [9] yet in the context of combined calibration of magnetometer and accelerometers. No attempt has been made so far in the previous literature to investigate the potential loss of those popular magnetometer calibration methods [4-8] by using the approximate ML instead of the optimal ML.

The main contribution of this paper is bringing to light the shortcomings of the popular three-axis magnetometer calibration methods founded on the approximate ML estimation and proposing to use the optimal ML estimation instead. The conclusion is potentially applicable to various kinds of three-axis sensors. The paper is organized as follows. Section II presents the magnetometer calibration problem in two forms of ML estimation: the optimal ML estimation and the approximate ML estimation, highlighting their relations and different statistical properties. Section III solves these two estimations using the Gauss-Newton method and Section IV compares their calibration performances by using synthetic simulation and real test data. The conclusions are given in Section V.

## II. Calibration Problem Formulation

Taking the time-invariant magnetic disturbance and sensor imperfection into account, the magnetometer measurement can be collectively modelled by [4, 6]

$$\mathbf{y} = \mathbf{S}\mathbf{C}_n^b \mathbf{m}^n + \mathbf{h} + \mathbf{e} \quad (1)$$

where $\mathbf{m}^n$ is a local magnetic vector in the local level frame (n-frame), $\mathbf{h}$ is the hard iron effect, $\mathbf{S}$ is the soft iron effect and $\mathbf{e}$ is i.i.d Gaussian noise with covariance $\sigma^2 \mathbf{I}_3$. The attitude matrix $\mathbf{C}_n^b$ transforms the geomagnetic vector from the local level frame to the magnetometer's body frame (b-frame). In a homogeneous external magnetic field like the geomagnetic field, $\mathbf{m}^n$ is constant and assumed to have unity norm without loss of generality. The model (1) is a rather general linear transformation that distorts and translates an unit sphere surface into an ellipsoid surface [4] and applies to many kinds of three-axis sensors like gyroscopes and accelerometers [2, 13].

The purpose of magnetometer calibration is to estimate the parameters $\mathbf{S}$ and $\mathbf{h}$ in the model (1). The magnetometer calibration problem can be formulated as an optimal maximum likelihood (ML) estimation [4, 9]

$$\min_{\boldsymbol{\theta}_{ml}} \sum_{k=1}^{N} \|\mathbf{e}_k\|^2 = \min_{\boldsymbol{\theta}_{ml}} \sum_{k=1}^{N} \|\mathbf{y}_k - \mathbf{S}\mathbf{C}_{n,k}^b \mathbf{m}^n - \mathbf{h}\|^2 \quad (2)$$
$$s.t. \quad \|\mathbf{m}^n\| = 1, \quad \mathbf{C}_{n,k}^b \in SO(3), \quad k = 1, \ldots, N$$

with variables $\boldsymbol{\theta}_{ml} = \{\mathbf{S}, \mathbf{h}, \mathbf{m}^n, \mathbf{C}_{n,k}^b\}$, where $\mathbf{C}_{n,k}^b$ is the magnetometer orientation for $k$-th data sample. It is not difficult to check that $\boldsymbol{\theta}_{ml}$ is not unique. For example, if $\mathbf{S}$ and $\mathbf{C}_{n,k}^b$ belong to one solution, then $\mathbf{S}\mathbf{Q}^T$ and $\mathbf{Q}\mathbf{C}_{n,k}^b$ would form another solution with any orthogonal matrix $\mathbf{Q}$. To get an unique solution, we should impose some constraints to the parameter $\boldsymbol{\theta}_{ml}$. Assume $\mathbf{S}^{-1} = \mathbf{Q}\mathbf{R}$ by the orthogonal-triangular (QR) decomposition, where $\mathbf{Q}$ is orthogonal and $\mathbf{R}$ is upper triangular with positive diagonal entries [17]. The second item in the squared objective function of (2)

$$\mathbf{S}\mathbf{C}_{n,k}^b \mathbf{m}^n = \mathbf{R}^{-1}\mathbf{Q}^T \mathbf{C}_{n,k}^b \mathbf{m}^n \triangleq \mathbf{R}^{-1}\mathbf{m}_k^{b^*} \quad (3)$$

where $\mathbf{m}_k^{b^*}$ also has unity norm as an orthogonal matrix keeps the vector length. The $b^*$-frame is implicitly defined according to the physical layout of the magnetic sensitive axes of the magnetometer, misaligning the above b-frame by the orthogonal matrix $\mathbf{Q}$. Specifically, the $b^*$-frame has its z-axis aligned with the z-sensor, y-axis orthogonal to z-axis in the plane formed by z-sensor and y-sensor, and x-axis naturally defined by the right-handed rule. Hereafter, the asterisk superscript will not be explicitly used for notational brevity. The three magnetic sensors' non-orthogonal matrix and scale factor matrix can be extracted by decomposing $\mathbf{R} = \mathbf{M}\boldsymbol{\Lambda}$, where $\boldsymbol{\Lambda}$ is a diagonal matrix making the diagonal of $\mathbf{M}$ be all ones.

The inverse of an upper triangular matrix is upper triangular as well. Denote $\mathbf{R}^{-1} \triangleq \mathbf{T}$, the ML estimation (2) is equivalently posed as

$$\min_{\boldsymbol{\theta}_{ml}} \sum_{k=1}^{N} \|\mathbf{e}_k\|^2 = \min_{\boldsymbol{\theta}_{ml}} \sum_{k=1}^{N} \|\mathbf{y}_k - \mathbf{T}\mathbf{m}_k^b - \mathbf{h}\|^2 \quad (4)$$
$$s.t. \quad \|\mathbf{m}_k^b\| = 1, \quad \mathbf{T} \in U(3), \quad k = 1, \ldots, N$$

with variables $\boldsymbol{\theta}_{ml} = \{\mathbf{T}, \mathbf{h}, \mathbf{m}_k^b\}$. $U(3)$ denotes the set of 3-by-3 upper triangular matrices. Now that the ML estimation (4) has an unique solution and the parameter space dimension of $\boldsymbol{\theta}_{ml}$ is $2N + 9$. As far as the magnetometer calibration is concerned, $\mathbf{m}_k^b (k = 1, \ldots, N)$ is a set of auxiliary constrained parameters of dimension $2N$. The formulation (4) is different from that in [4] (Eq. (6) therein) which used a product of an orthogonal matrix and a diagonal matrix in place of $\mathbf{T}$. From the algorithmic viewpoint, (4) is more preferable as $\mathbf{T}$ only contains six unconstrained entries, while special care has to be taken to handle the orthogonal matrix (with three freedom) in [4]'s formulation.

Alternatively, we can reduce the original calibration problem (1) to a suboptimal estimation of considerably smaller dimension, by removing the auxiliary parameters $\mathbf{m}_k^b$ with unity norm. Using (1)

and QR decomposition of $\mathbf{S}$, we have

$$\begin{aligned}1 = \|\mathbf{m}^n\|^2 &= \|\mathbf{C}_b^n \mathbf{S}^{-1}(\mathbf{y}-\mathbf{h}-\mathbf{e})\|^2 = \|\mathbf{R}(\mathbf{y}-\mathbf{h}-\mathbf{e})\|^2 \\ &= \|\mathbf{R}(\mathbf{y}-\mathbf{h})\|^2 - 2(\mathbf{y}-\mathbf{h})^T \mathbf{R}^T \mathbf{R}\mathbf{e} + \mathbf{e}\mathbf{R}^T\mathbf{R}\mathbf{e} \\ &\triangleq \|\mathbf{R}(\mathbf{y}-\mathbf{h})\|^2 + w \end{aligned} \quad (5)$$

where the defined noise $w$ is exactly not Gaussian as it contains a quadratic item of $\mathbf{e}$ and we have

$$\begin{aligned} \mu_w &\triangleq E(w) = E\left[tr\left(\mathbf{R}^T \mathbf{R}\mathbf{e}\mathbf{e}^T\right)\right] = tr\left(\mathbf{R}^T\mathbf{R}\right)\sigma^2 > 0 \\ \sigma_w^2 &\triangleq E(w^2) - E^2(w) \end{aligned} \quad (6)$$

Approximating $w$ by a Gaussian, i.e., $w \sim \mathbb{N}(\mu_w, \sigma_w^2)$, then an approximate ML formulation can be obtained as [7]

$$\min_{\boldsymbol{\theta}_{nm}} \sum_{k=1}^{N} \left\{ \frac{1}{\sigma_w^2}\left[1 - \|\mathbf{R}(\mathbf{y}_k-\mathbf{h})\|^2 - \mu_w\right]^2 + \log \sigma_w^2 \right\} \quad (7)$$
$$s.t. \quad \mathbf{R} \in U(3)$$

with variables $\boldsymbol{\theta}_{nm} = \{\mathbf{R}, \mathbf{h}\}$. Mean $\mu_w$ and variance $\sigma_w^2$ occur in the objective function as they both depend on the calibration parameter. This objective function is more complex than that in (4), so for simplicity many previous works just discard the items of $\mu_w$ and $\sigma_w^2$, leading to such a suboptimal estimation as

$$\min_{\boldsymbol{\theta}_{nm}} \sum_{k=1}^{N} \left[1 - \|\mathbf{R}(\mathbf{y}_k-\mathbf{h})\|^2\right]^2 \quad (8)$$
$$s.t. \quad \mathbf{R} \in U(3)$$

The parameter space dimension of $\boldsymbol{\theta}_{nm}$ is reduced to as small as 9. The parameters $\mathbf{m}_k^b$ have been completely condensed by the norm operation, as denoted by the subscript '$nm$'. Hereafter it is referred to as the NM estimation to distinguish from the optimal ML estimation (4). The above subjective function is quartic in parameters. It complicates the calibration problem with multiple minima and maxima, so we have to make sure find a good initial estimate for the nonlinear solver in next section. The formulation (8) is the basis of various attitude-independent magnetometer calibration methods in the literature, e.g., [3-6, 8, 16], with different constraints on the matrix $\mathbf{R}$. For instance, it is symmetric in [8] and a general matrix followed by a singular value decomposition in [4].

We now see that the parameter $\mathbf{S}$ can only be determined up to an orthogonal matrix in either ML or NM estimations. With the obtained parameters $\mathbf{R}$ (or $\mathbf{T}$) and $\mathbf{h}$, the calibrated magnetometer measurement can be expressed in the physically defined b-frame as

$$\mathbf{m}_k^b = \mathbf{R}(\mathbf{y}_k - \mathbf{h}) = \mathbf{T}^{-1}(\mathbf{y}_k - \mathbf{h}) \quad (9)$$

## III. ESTIMATION IMPLEMENTATION

We need to numerically solve two nonlinear minimizations for the optimal ML estimation (4) and the suboptimal NM estimation (8), respectively. Good initial estimates are available for both minimizations, so the efficient Gauss-Newton method [18] is adopted herein. Given the analytic Jacobian and Hessian information, the Gauss-Newton method updates the estimate as such

$$\mathbf{x}^{(i+1)} = \mathbf{x}^{(i)} - \left[\nabla_\mathbf{x}^2 f\big|_{\mathbf{x}^{(i)}}\right]^{-1} \left(\nabla_\mathbf{x} f\big|_{\mathbf{x}^{(i)}}\right), \quad i = 0, 1, \ldots \quad (10)$$

where $\nabla_\mathbf{x} f$ is the Jacobian vector and $\nabla_\mathbf{x}^2 f$ is the Hessian matrix of the objective with respect to the estimate $\mathbf{x}$.

### A. Suboptimal NM Estimate

For the NM estimation (8), the minimization objective $f = \sum_{k=1}^{N}\left[1-\|\mathbf{R}(\mathbf{y}_k-\mathbf{h})\|^2\right]^2$. Let $\mathbf{u}_k \triangleq \mathbf{y}_k - \mathbf{h}$, then

$$f = \sum_{k=1}^{N}\left[1-\|\mathbf{R}\mathbf{u}_k\|^2\right]^2 \quad (11)$$

The estimate here is defined as $\mathbf{x} = \left[vec^T(\mathbf{R}) \quad \mathbf{h}^T\right]^T$, where $vec(\mathbf{R})$ forms a vector by stacking the columns of the matrix $\mathbf{R}$ but excluding the lower triangular zero entries. The Jacobian vector and Hessian matrix can be respectively derived as

$$\nabla_\mathbf{x} f_* = 4\sum_{k=1}^{N}\left[\mathbf{J}_R^T \quad \mathbf{J}_h^T\right]^T \quad (12)$$

$$\nabla_\mathbf{x}^2 f_* = 4\sum_{k=1}^{N}\begin{bmatrix}\mathbf{H}_{RR} & \mathbf{H}_{Rh} \\ \mathbf{H}_{Rh}^T & \mathbf{H}_{hh}\end{bmatrix}_{9\times 9}^T \quad (13)$$

where

$$\begin{aligned} \mathbf{J}_R &= \left(\|\mathbf{R}\mathbf{u}_k\|^2 - 1\right)\mathbf{u}_k \otimes (\mathbf{R}\mathbf{u}_k) \\ \mathbf{J}_h &= -\left(\|\mathbf{R}\mathbf{u}_k\|^2 - 1\right)\mathbf{R}^T\mathbf{R}\mathbf{u}_k \end{aligned} \quad (14)$$

and

$$\begin{aligned} \mathbf{H}_{RR} &= 2\left(\mathbf{u}_k\mathbf{u}_k^T\right)\otimes\left(\mathbf{R}\mathbf{u}_k\mathbf{u}_k^T\mathbf{R}^T\right) + \left(\|\mathbf{R}\mathbf{u}_k\|^2 - 1\right)\left(\mathbf{u}_k\mathbf{u}_k^T\right)\otimes \mathbf{I} \\ \mathbf{H}_{Rh} &= -2\left(\mathbf{u}_k\otimes(\mathbf{R}\mathbf{u}_k)\right)\mathbf{u}_k^T\mathbf{R}^T\mathbf{R} - \left(\|\mathbf{R}\mathbf{u}_k\|^2 - 1\right)\left(\mathbf{I}\otimes(\mathbf{R}\mathbf{u}_k) + \mathbf{u}_k\otimes\mathbf{R}\right) \\ \mathbf{H}_{hh} &= \left(\|\mathbf{R}\mathbf{u}_k\|^2 - 1\right)\mathbf{R}^T\mathbf{R} + 2\mathbf{R}^T\mathbf{R}\mathbf{u}_k\mathbf{u}_k^T\mathbf{R}^T\mathbf{R} \end{aligned}$$

(15)

where the operator $\otimes$ denotes the Kronecker product. Asterisk subscripts in $\nabla_\mathbf{x} f_*$ and $\nabla_\mathbf{x}^2 f_*$ mean that the columns and/or rows corresponding to the excluded lower triangular entries have been removed.

### B. Optimal ML Estimate

For the optimal ML estimation (4), the minimization objective $f = \sum_{k=1}^{N}\|\mathbf{y}_k - \mathbf{T}\mathbf{m}_k^b - \mathbf{h}\|^2 + \lambda_k\left(\|\mathbf{m}_k^b\|^2 - 1\right)$ where $\lambda_k$ is the

Lagrange coefficient for the unity norm constraint of $\mathbf{m}_k^b$. The estimate in this case becomes $\mathbf{x} = \begin{bmatrix} vec^T(\mathbf{T}) & \mathbf{h}^T & \mathbf{m}_1^b \ldots \mathbf{m}_N^b & \lambda_1 \ldots \lambda_N \end{bmatrix}^T$. The dimension of estimate expands (from $2N+9$ in (4)) to $4N+9$ in implementation. The Jacobian vector and Hessian matrix can be respectively derived as

$$\nabla_{\mathbf{x}} f = \begin{bmatrix} \mathbf{J}_T^T & \mathbf{J}_h^T & \mathbf{J}_{m_k^b}^T & \mathbf{J}_{\lambda_k}^T \\ & & k=1:N & k=1:N \end{bmatrix}^T \quad (16)$$

$$\nabla_{\mathbf{x}}^2 f = \begin{bmatrix} \mathbf{H}_{TT} & \mathbf{H}_{Th} & \mathbf{H}_{Tm_k^b} \cdots & \mathbf{0}_{9\times 1} \cdots \\ \mathbf{H}_{Th}^T & \mathbf{H}_{hh} & \mathbf{H}_{hm_k^b} \cdots & \mathbf{0}_{3\times 1} \cdots \\ \mathbf{H}_{Tm_k^b}^T & \mathbf{H}_{hm_k^b}^T & \mathbf{H}_{m_k^b m_k^b} \cdots & \mathbf{H}_{m_k^b \lambda_k} \cdots \\ \vdots & \vdots & \vdots & \vdots \\ \mathbf{0}_{9\times 1}^T & \mathbf{0}_{3\times 1}^T & \mathbf{H}_{m_k^b \lambda_k}^T \cdots & 0 \cdots \\ \vdots & \vdots & \vdots & \vdots \\ & & \underbrace{\phantom{XX}}_{k=1:N} & \underbrace{\phantom{XX}}_{k=1:N} \end{bmatrix}_{(4N+9)\times(4N+9)} \quad (17)$$

where

$$\mathbf{J}_T = -2\sum_{k=1}^{N} \mathbf{m}_k^b \otimes (\mathbf{u}_k - \mathbf{T}\mathbf{m}_k^b), \quad \mathbf{J}_h = -2\sum_{k=1}^{N}(\mathbf{u}_k - \mathbf{T}\mathbf{m}_k^b) \quad (18)$$
$$\mathbf{J}_{m_k^b} = -2\mathbf{T}^T(\mathbf{u}_k - \mathbf{T}\mathbf{m}_k^b) + 2\lambda_k \mathbf{m}_k^b, \quad \mathbf{J}_{\lambda_k} = \|\mathbf{m}_k^b\|^2 - 1$$

and

$$\mathbf{H}_{TT} = 2\sum_{k=1}^{N}(\mathbf{m}_k^b \mathbf{m}_k^{bT}) \otimes \mathbf{I}, \quad \mathbf{H}_{Th} = 2\sum_{k=1}^{N}\mathbf{m}_k^b \otimes \mathbf{I}$$
$$\mathbf{H}_{Tm_k^b} = 2\big((\mathbf{m}_k^b \otimes \mathbf{I})\mathbf{T} - \mathbf{I} \otimes (\mathbf{u}_k - \mathbf{T}\mathbf{m}_k^b)\big) \quad (19)$$
$$\mathbf{H}_{hh} = 2N\mathbf{I}, \quad \mathbf{H}_{hm_k^b} = 2\mathbf{T}$$
$$\mathbf{H}_{m_k^b m_k^b} = 2\mathbf{T}^T\mathbf{T} + 2\lambda_k \mathbf{I}, \quad \mathbf{H}_{m_k^b \lambda_k} = 2\mathbf{m}_k^b$$

The matrix equality $vec(\mathbf{ABC}) = (\mathbf{C}^T \otimes \mathbf{A}) vec(\mathbf{B})$ has been frequently used in deriving (14), (15), (18) and (19).

*C. Initial Estimate*

The above two ML estimations both require batch-processing in nature. A good initial estimate can be derived from (5), as given in [9] or [13, 14]. The minimum objective at the true estimate should be close to zero, so it is reasonable to consider $1 = \|\mathbf{R}(\mathbf{y}_k - \mathbf{h})\|^2$ to find an initial estimate. Expanding the expression,

$$\mathbf{y}_k^T \mathbf{A} \mathbf{y}_k + \mathbf{b}^T \mathbf{y}_k + c = 0 \quad (20)$$

where $\mathbf{A} = \mathbf{R}^T\mathbf{R}$, $\mathbf{b} = -2\mathbf{R}^T\mathbf{R}\mathbf{h}$ and $c = \mathbf{h}^T\mathbf{R}^T\mathbf{R}\mathbf{h} - 1$. The equation can be written as a linear equation of unknowns as

$$\begin{bmatrix} \mathbf{y}_k^T \otimes \mathbf{y}_k^T & \mathbf{y}_k^T & 1 \end{bmatrix} \begin{bmatrix} vec(\mathbf{A}) \\ \mathbf{b} \\ c \end{bmatrix} \triangleq \mathbf{Y}_k \mathbf{z} = 0, \quad k=1,\ldots,N \quad (21)$$

or collectively,

$$\mathbf{Y}\mathbf{z} = 0 \quad (22)$$

with $\mathbf{Y} = \begin{bmatrix} \mathbf{Y}_1^T & \ldots & \mathbf{Y}_N^T \end{bmatrix}^T$. As $\mathbf{A}$ is symmetric, $vec(\mathbf{A})$ is formed by stacking the columns of the matrix $\mathbf{A}$ but excluding the lower triangular entries. The columns of $\mathbf{Y}$ corresponding to the three lower triangular entries are merged to those columns corresponding to their symmetric counterparts.

Regarding (22) as a linear least-square problem, i.e., $\mathbf{z} = \min \|\mathbf{Y}\mathbf{z}\|^2$. Its solution should satisfy the normal equation of least squares, $\mathbf{Y}^T\mathbf{Y}\mathbf{z} = 0 = 0 \cdot \mathbf{z}$. That is to say, the solution should be the eigenvector of $\mathbf{Y}^T\mathbf{Y}$ with zero (or minimum) eigenvalue, as a non-negative symmetric matrix has non-negative eigenvalues. Denote this solution as $\mathbf{z}_e$. Noticing that $\alpha \mathbf{z}_e$ for any real $\alpha$ is also a solution to (22), we assume

$$\begin{bmatrix} vec(\mathbf{A}) \\ \mathbf{b} \\ c \end{bmatrix} = \alpha \mathbf{z}_e = \alpha \begin{bmatrix} vec(\mathbf{A}_e) \\ \mathbf{b}_e \\ c_e \end{bmatrix} \quad (23)$$

From (20), $1 = \mathbf{h}^T\mathbf{R}^T\mathbf{R}\mathbf{h} - c = \mathbf{b}^T\mathbf{A}^{-1}\mathbf{b}/4 - c = \alpha(\mathbf{b}_e^T\mathbf{A}_e^{-1}\mathbf{b}_e/4 - c_e)$, so $\alpha = 4/(\mathbf{b}_e^T\mathbf{A}_e^{-1}\mathbf{b}_e - 4c_e)$.

Then $\mathbf{h}^{(0)} = -\mathbf{A}^{-1}\mathbf{b}/2$ and $\mathbf{R}^{(0)} = chol(\mathbf{A})$, where $chol(\cdot)$ denotes the matrix Cholesky factorization. For the optimal ML estimation, $\mathbf{T}^{(0)} = (\mathbf{R}^{(0)})^{-1}$, the initial Lagrange coefficient $\lambda_k^{(0)} = 0$ and the initial $\mathbf{m}_k^{b(0)}$ can be readily obtained by (9).

IV. SIMULATION AND TEST RESULTS

We first carry out a synthetic magnetometer calibration to examine and compare the above two estimations. The vector $\mathbf{m}^n$ is taken to be the geomagnetic field unit vector in the Changsha city, $\mathbf{m}^n = \begin{bmatrix} 0.7388 & 0.0409 & -0.6727 \end{bmatrix}^T$ (North, Upward, East) according to the World Magnetic Model 2005. Hereafter the magnetic field units are Gauss, if not explicitly stated. The true soft and hard effects in the measurement model (1) are taken to be

$$\mathbf{S} = \begin{bmatrix} 0.7 & -0.8 & 0.4 \\ 1.1 & 0.3 & -0.1 \\ -0.3 & 0.6 & 0.7 \end{bmatrix}, \quad \mathbf{h} = \begin{bmatrix} 0.5 \\ 1.7 \\ 2.6 \end{bmatrix} \quad (24)$$

The standard deviation of the measurement noise $\sigma = 0.003$. The attitude matrix $\mathbf{C}_n^b$ is re-parameterized in Euler angles

$$\mathbf{C}_n^b = \begin{bmatrix} \cos\theta\cos\psi & \sin\phi\sin\psi - \cos\phi\cos\psi\sin\theta & \cos\phi\sin\psi + \cos\psi\sin\phi\sin\theta \\ \sin\theta & \cos\phi\cos\theta & -\cos\theta\sin\phi \\ -\cos\theta\sin\psi & \cos\psi\sin\phi + \cos\phi\sin\theta\sin\psi & \cos\phi\cos\psi - \sin\phi\sin\theta\sin\psi \end{bmatrix}$$

with the three angles given as functions of the data index ($k = 1, \ldots, N$) as

$$\phi = 20\sin(20\pi k/N + \pi/2)$$
$$\psi = 360k/N \quad \text{(unit: degree)} \quad (26)$$
$$\theta = 20\sin(20\pi f/N)$$

To facilitate the performance evaluation, the error metrics in [6] are adapted in this paper. According to the matrix decompositions above in Section II, $\mathbf{S} = \mathbf{R}^{-1}\mathbf{Q}^T = \mathbf{\Lambda}^{-1}\mathbf{M}^{-1}\mathbf{Q}^T$. Several physical error metrics are defined, where the average scale factor error $e_s = \frac{1}{3}\|diag(\mathbf{\Lambda}^{-1}\hat{\mathbf{\Lambda}} - \mathbf{I})\| \times 100\%$ (in percentage), the average sensor orthogonal error $e_o = \frac{180}{3\pi} \cdot \|vec(\hat{\mathbf{M}} - \mathbf{M})\|$ (in degree), and the average hard-iron effect error $e_h = \frac{1}{3}\|\hat{\mathbf{h}} - \mathbf{h}\|$ (in Gauss). The hatted quantities mean the final ML estimate or NM estimate.

Figure 1 plots the data points generated by the measurement model (1) when $N = 300$, along with the ellipsoid surface (left-upper corner) determined by the true parameters (24). Figure 2 gives the magnitudes of all data points $\mathbf{y}_k$ that substantially deviate from unity because of the distorting transformation. Figures 1-2 also plot the roughly calibrated data points $\mathbf{m}_k^{b(0)}$ using the initial estimate given in Sec. III.C. The fact $\mathbf{m}_k^{b(0)}$ approaching unity in magnitude manifests the goodness of the initial estimate.

We then implement 50 Monte Carlo runs for both estimations. Their objective function values for each iteration are presented in Fig. 3, namely, (4) for the ML estimation and (8) for the NM estimation. Both estimations converge well within five iterations. The NM initial objective values at 0th iteration are close to the minimum, confirming the initial estimates are very good. Note that the ML initial objective values are roughly zero just because the auxiliary parameter is initially determined by (9). Figure 4 presents a boxplot across 50 Monte Carlo runs for the above three error metrics ($e_s$, $e_o$ and $e_h$) of both estimations in pairs. The last metric $e_h$ is scaled up by 300 for better presentation. Each box has lines at the lower quartile, median, and upper quartile values. Whiskers extend from each end of the box to the adjacent values in the data; the most extreme values within 1.5 times the interquartile range from the ends of the box. Outliers are data with values beyond the ends of the whiskers and displayed with a red + sign. Their corresponding means and standard deviations are listed in Table I. The ML estimation performs slightly better than the NM estimation in both mean and standard deviation. The execution time of both estimations by Matlab is compared in Figure 5 for different $N$. Expectedly, the ML estimation increases quickly in execution time along with the number of data samples. When $N = 1000$ for instance, its computation cost is about ten times that of the NM estimation.

Further Monte Carlo runs are made to examine the estimation sensitivity to initial estimate error. We randomly change the initial estimate by several percentage and record the number of divergence runs. In specific, the changed initial estimate is $\tilde{\mathbf{R}}^{(0)} = \mathbf{R}^{(0)} \cdot {}^* [\mathbf{1}_{3\times 3} + \alpha \cdot sign(randn(3,3))]$ and $\tilde{\mathbf{h}}^{(0)} = \mathbf{h}^{(0)} \cdot {}^* [\mathbf{1}_{3\times 1} + \alpha \cdot sign(randn(3,1))]$, where $\alpha$ is a varying percentage, .* the element-by-element product between matrices, $\mathbf{1}$ a matrix of all ones of appropriate dimensions, and $sign(\cdot)$ and $randn(\cdot)$ are the sign function and the normally distributed random number function, respectively. Figure 6 presents the number of divergence out of 50 runs as the initial estimate is varied within 7%. Referring to Fig. 3, we consider those runs as divergence for which the final objective function values are larger than the thresholds of 0.018 (NM) or 0.004 (ML). The NM estimation starts to diverge when the initial estimate is changed by as small as 1%, while the ML estimation does not diverge within 5% change. When the initial estimate is changed by 3% or more, the NM estimation diverges in over a half runs and completely crashes by 6%. It can be concluded that the NM estimation has a much narrower convergence attractive region and is far more sensitive to initial error than the ML estimation does, which is owed to its quartic attribute. Once again, it shows the algorithm in Sec. III.C is a quite good initial estimate for the Gauss-Newton method.

Two magnetometer datasets were collected using an Xsens MTi-G-700 unit in an open area. The raw measurements are plotted in Fig. 7, along with their ellipsoid surfaces determined by the respective calibration results listed in Table II. The first dataset points cover most part of the left ellipsoid surface, while the second dataset points concentrate only at the bottom of the right ellipsoid surface. In other words, the first dataset carries relatively richer information about the ellipsoid surface and is supposed to give a better calibration result. As seen in Table II, the NM and ML estimations yield identical results for the first dataset, yet showing discrepancy for the second dataset. The ML estimate for the second dataset seems better than that of NM as the former is closer to the result of the first dataset. This is confirmed by Fig. 8 that plots the magnitude and error histogram of calibrated measurements of the first dataset when respectively applying the ML and NM estimates from the second dataset, and by Table III that lists the error metrics of the estimates from the second dataset when we take the estimates from the first dataset as the reference. This evidences that the ML estimation is able to yield consistently good result even for datasets with insufficient attitude maneuvers. This property is particularly beneficial to land applications where the vehicle motion is usually confined to a flat surface. Finally, their sensitivities to initial errors for the two test datasets are also examined by Monte Carlo runs. As summarized in Fig. 9, the ML estimation does not diverge within 25% and 4% initial estimate change, respectively in the first and second datasets, demonstrating considerably larger convergence region than the NM estimation. This agrees with our previous observation in simulations. Additionally, the first dataset obviously tolerates larger initial errors for both estimations than the second dataset does, owed to richer attitude maneuvers as shown in Fig. 7.

## V. CONCLUSIONS

A three-axis magnetometer has wide applications in attitude determination and scientific measurement. Due to the compound effect of sensor imperfection and vulnerability to ambient magnetic disturbances, the three-axis magnetometer needs to be carefully calibrated prior to any practical use. Attitude-independent methods have been most popular for magnetometer calibration. These methods make use of the constant magnitude relationship in a homogeneous magnetic field to accomplish the calibration by way of ML estimation. This paper throws lights on the approximate and quartic characteristics of previous ML methods and proposes to use the quadratic optimal ML estimation for magnetometer calibration. The two ML calibrations are extensively compared using magnetometer simulations and test datasets. The optimal ML calibration outperforms the popular approximate ML method for magnetometer calibration in accuracy and stability, especially for those situations with insufficient attitude maneuvers. The approximate ML method's higher sensitivity to initial errors would potentially lead to magnetometer calibration failure in cases where a fine initial estimate was unavailable. Although the optimal ML calibration is relatively computation-intensive, it is out of problem for magnetometer calibration which is often an offline process. In view of the generality of the measurement model discussed in this paper, the conclusions obtained naturally apply to many kinds of three-axis sensors, including but not limited to inertial sensors like gyroscopes and accelerometers.

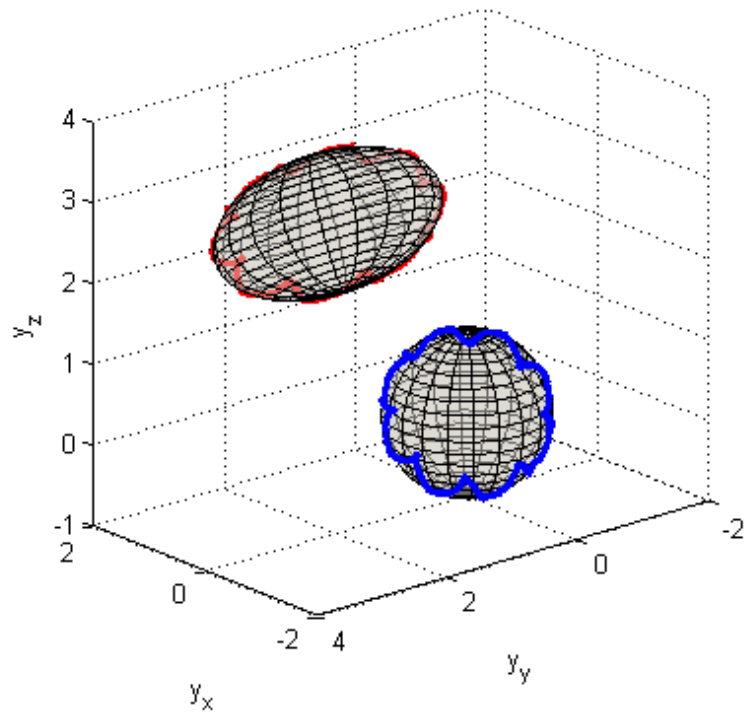

Figure 1. Data points, before (red dots, left-upper) and after (blue dots, right-lower) applying initial estimate.

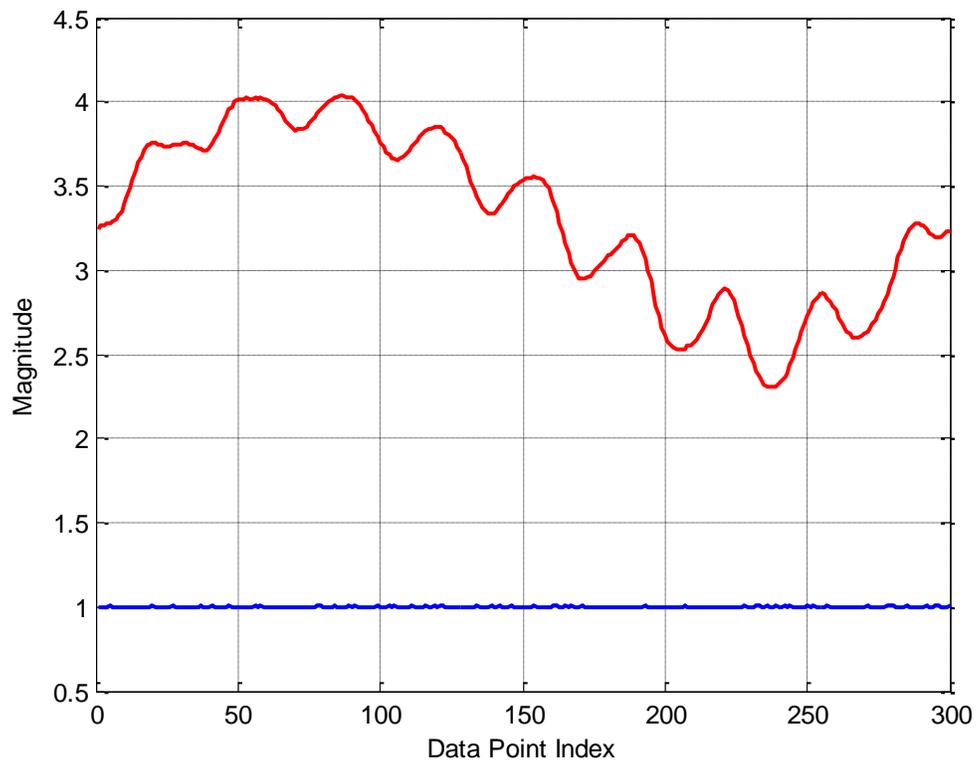

Figure 2. Magnitude of data points, before (red line) and after (blue line) applying initial calibration.

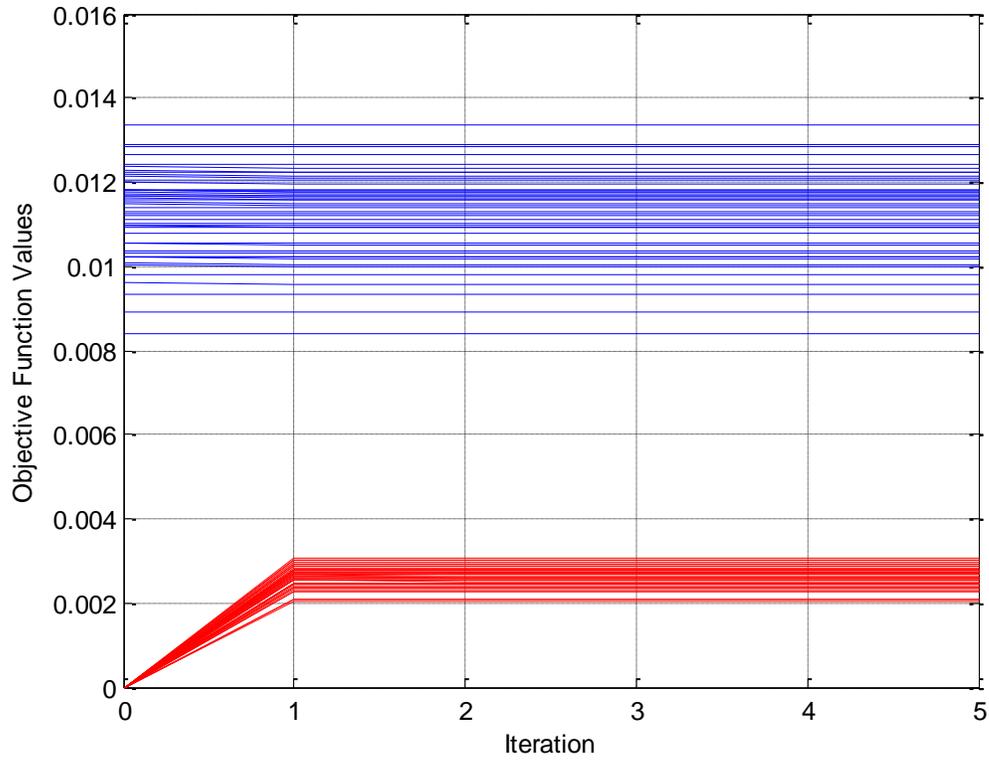

Figure 3. Objective function values at each iteration across 50 Monte Carlo runs (NM: blue line; ML: red line)

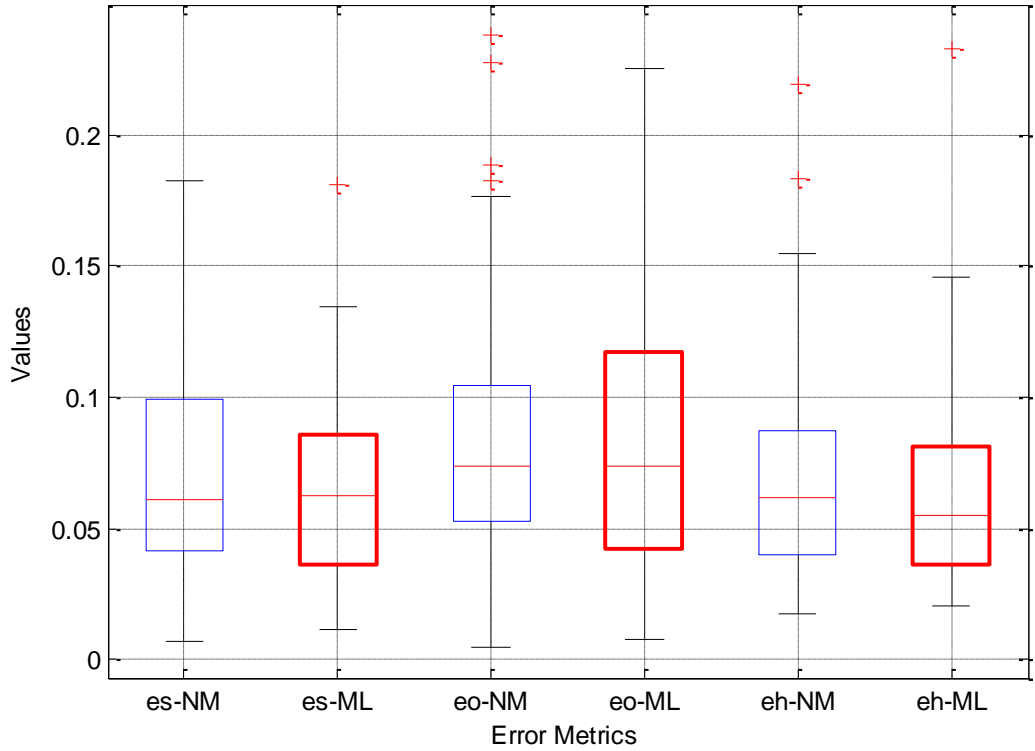

Figure 4. Boxplot for three error metrics $e_s$, $e_o$ and scaled $e_h$ (NM: blue box; ML: red box).

Table I. Mean (Standard Deviation) of Three Error Metrics

|    | $e_s$ (%)         | $e_o$ (deg)       | $e_h$ (Gauss)     |
|----|-------------------|-------------------|-------------------|
| **NM** | 0.0704 (0.0405) | 0.0864 (0.0529) | 0.0002 (0.0001) |
| **ML** | 0.0662 (0.0361) | 0.0799 (0.0467) | 0.0002 (0.0001) |

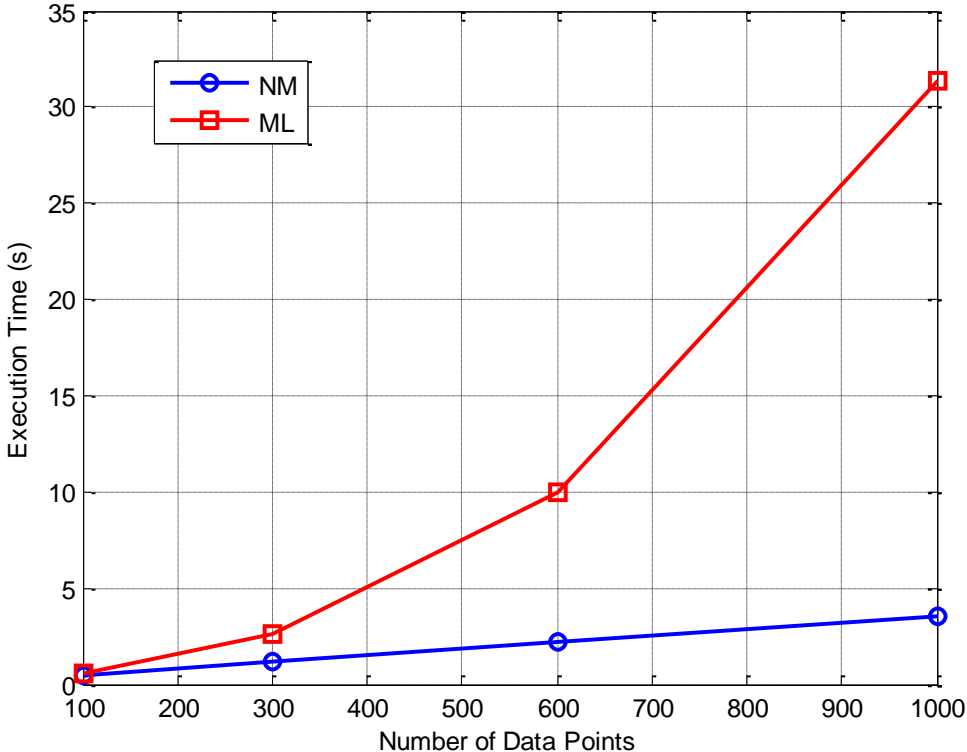

Figure 5. Execution time comparison for different number of data points (NM: blue line; ML: red line).

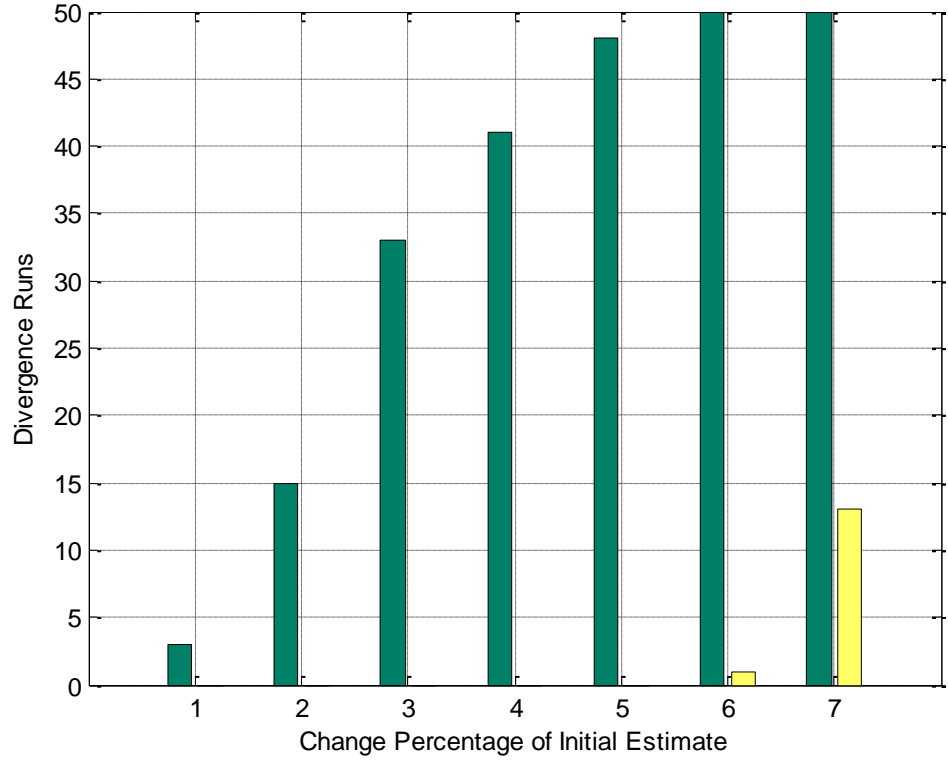

Figure 6. Number of divergence out of 50 runs as initial estimate varied by a range of percentages in simulations (NM: green bar; ML: yellow bar).

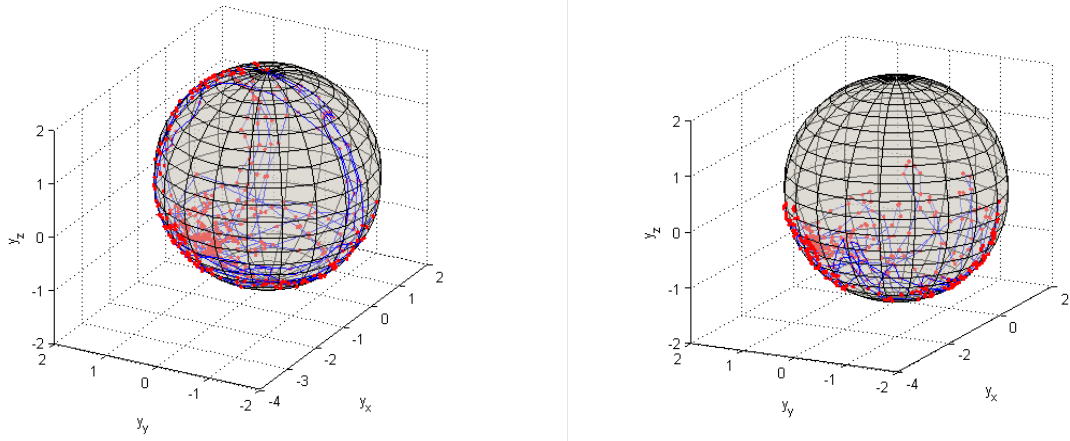

Figure 7. Two raw magnetometer datasets by Xsens MTi-G-700 and their fitted ellipoids.

Table II. Calibration Results of Two Datasets

|  | Dataset #1 | Dataset #2 |
|---|---|---|
| **NM** | $R = \begin{bmatrix} 0.5093 & -0.0018 & 0.0021 \\ 0 & 0.5104 & -0.0006 \\ 0 & 0 & 0.5115 \end{bmatrix}$ | $R = \begin{bmatrix} 0.5070 & -0.0014 & -0.0003 \\ 0 & 0.5073 & 0.0007 \\ 0 & 0 & 0.4994 \end{bmatrix}$ |
|  | $d = \begin{bmatrix} -0.0621 & -0.0036 & 0.0247 \end{bmatrix}^T$ | $d = \begin{bmatrix} -0.0549 & -0.0050 & 0.0612 \end{bmatrix}^T$ |
| **ML** | $T^{-1} = \begin{bmatrix} 0.5093 & -0.0018 & 0.0021 \\ 0 & 0.5104 & -0.0006 \\ 0 & 0 & 0.5115 \end{bmatrix}$ | $T^{-1} = \begin{bmatrix} 0.5090 & -0.0012 & -0.0013 \\ 0 & 0.5099 & -0.0007 \\ 0 & 0 & 0.5077 \end{bmatrix}$ |
|  | $d = \begin{bmatrix} -0.0621 & -0.0036 & 0.0247 \end{bmatrix}^T$ | $d = \begin{bmatrix} -0.0525 & -0.0014 & 0.0311 \end{bmatrix}^T$ |

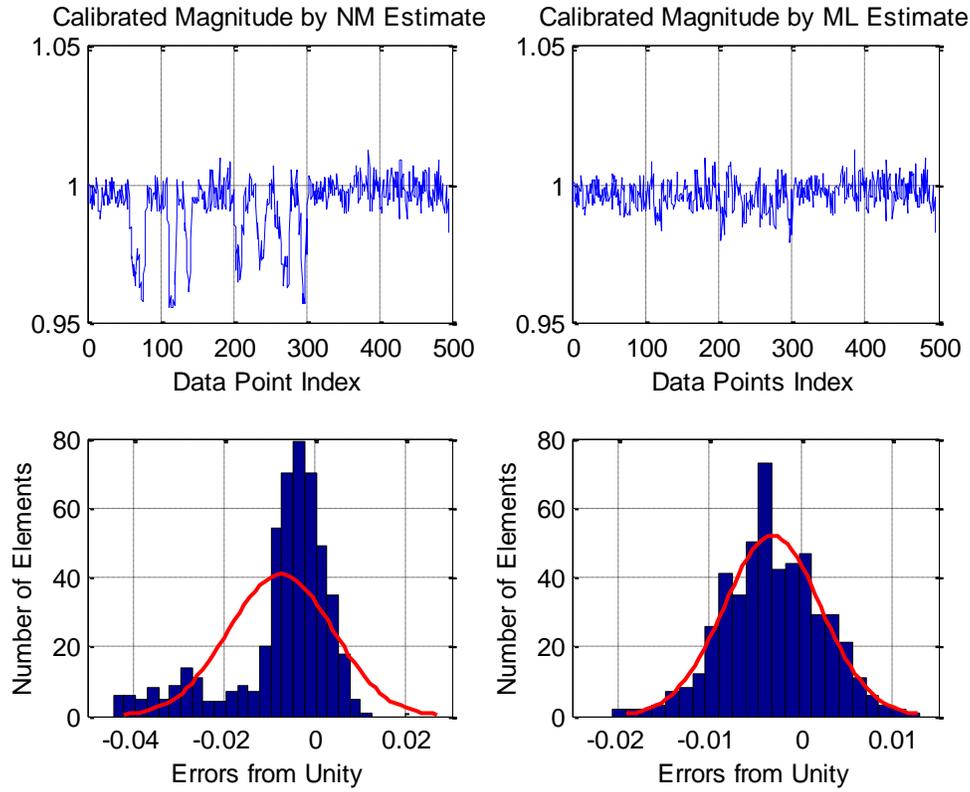

Figure 8. Magnitude and error histogram of calibrated measurements of dataset #1 when applying calibration parameters from dataset #2, with NM estimate (left column) and ML estimate (right column). Red curves in bottom two figures are fitted normal distributions from histograms.

Table III. Estimates Error Metrics from Dataset #2 (Referencing Estimates from Dataset #1)

|     | $e_s$ (%) | $e_o$ (deg) | $e_h$ (Gauss) |
|-----|-----------|-------------|---------------|
| **NM** | 0.8287 | 0.1026 | 0.0124 |
| **ML** | 0.2527 | 0.1281 | 0.0039 |

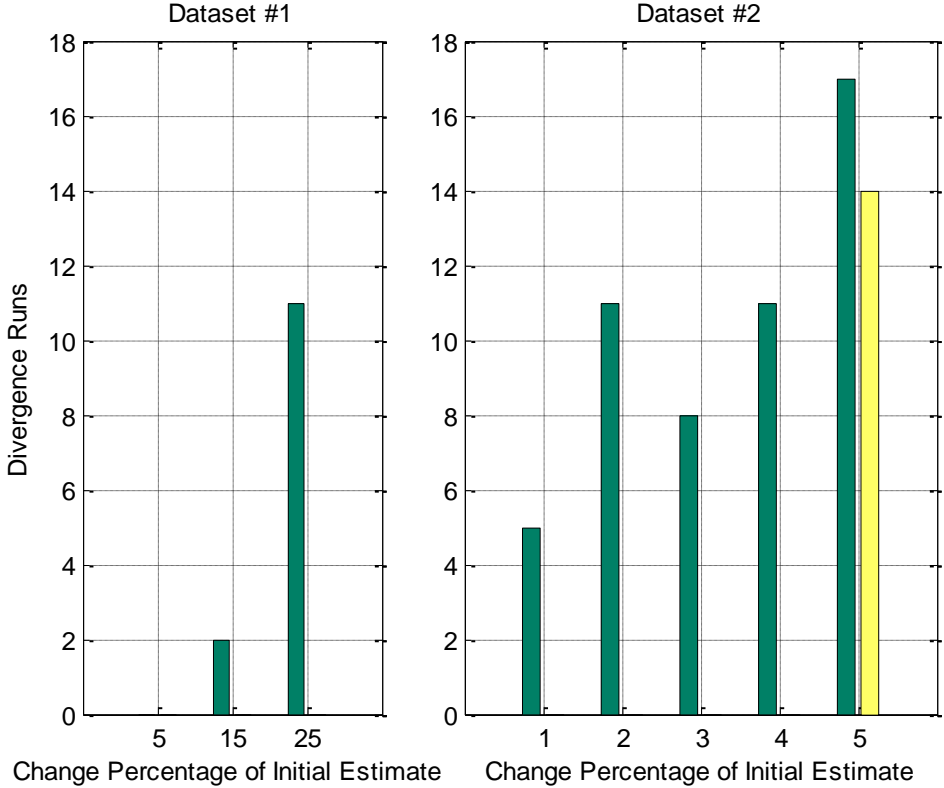

Figure 9. Number of divergence out of 50 runs as initial estimate varied by a range of percentages in tests (NM: green bar; ML: yellow bar).